\documentclass{article}

\usepackage{microtype}
\usepackage{graphicx}
\usepackage{subcaption}
\usepackage{booktabs} 

\usepackage{hyperref}

\usepackage[accepted]{icml2026}

\usepackage{amsmath}
\usepackage{amssymb}
\usepackage{mathtools}
\usepackage{amsthm}

\usepackage[capitalize,noabbrev]{cleveref}

\theoremstyle{plain}

\theoremstyle{definition}

\theoremstyle{remark}

\usepackage[textsize=tiny]{todonotes}

\icmltitlerunning{Reliability, Faithfulness, and the Limits of Post-hoc Explanations of Opaque Scientific Models}

\begin{document}

\twocolumn[
  \icmltitle{Reliability, Faithfulness, and the Limits of\\Post-hoc Explanations of Opaque Scientific Models}



  \icmlsetsymbol{equal}{*}

  \begin{icmlauthorlist}
    \icmlauthor{Nick Oh}{comp}
    \icmlauthor{Helen Jin}{yyy}
  \end{icmlauthorlist}

  \icmlaffiliation{yyy}{Department of Computer and Information Science, University of Pennsylvania, Philadelphia, Pennsylvania, United States}
  \icmlaffiliation{comp}{socius labs, London, United Kingdom}

  \icmlcorrespondingauthor{Nick Oh}{nick.sh.oh@socius.org}
  \icmlcorrespondingauthor{Helen Jin}{helenjin@engineering.upenn.edu}

  \icmlkeywords{Machine Learning, ICML}

  \vskip 0.3in
]



\printAffiliationsAndNotice{}  

\begin{abstract}
Post-hoc explanation methods are routinely used to interpret scientific machine learning models, with the deliverable understood to be insight into the phenomenon the model has been trained on. The transition may be taken to be secured once the model is reliable enough and the explanation faithful enough. We argue it is not. Reliability checks that the model's predictions match the phenomenon's outcomes, and faithfulness checks that the explanation matches the model, but neither checks whether the model works as the phenomenon works, which is what a claim about structure requires. The chain can support candidate hypotheses under external corroboration, but it cannot, on its own, support claims about how the phenomenon is in fact structured.
\end{abstract}

\section{Introduction}
Post-hoc explainable AI (XAI) methods --- SHAP, LIME, attention attribution, gradient-based saliency, Layer-wise Relevance Propagation (LRP), TCAV and Concept Relevance Propagation, counterfactual explanations, and related techniques --- are routinely deployed to interpret models built for scientific discovery and hypothesis generation. The practice now spans neuroscience \citep{ryali2024deep, goktepe2025does, hofmann2025utility, eilts2026explainable}, cognitive psychology \citep{auletta2023predicting, wang2024rethink, kumar2024shared}, physics \citep{vent2026physics, bonalumi2024explainable}, biochemistry \citep{pampari2025chrombpnet, mao2024phenotype, sun2025esm2_amp}, economics \citep{chung2023inside, yuan2024deep, goswami2026significance, ji2026multi}, medicine \citep{atad2022chexplaining, rahnfeld2025comparative, klein2026explainable}, the earth and environmental sciences \citep{stomberg2023recognizing, graciosa2025uncovering, vinokic2025effectiveness}, and sociology \citep{benslimane2024shap}. In each case the model functions not as a black-box predictor whose only credential is empirical accuracy, but as a putative source of insight into the phenomenon it has been trained on. The intended deliverable is understanding \textit{with} the model, rather than understanding \textit{of} the model \citep{sullivan2022understanding}.
 
The transition from understanding \textit{of} the model to understanding \textit{about} the world is taken to be secured by two materials whose standards are themselves the subject of extensive methodological work \citep{hooker2019benchmark, laugel2019dangers, laugel2019issues, slack2020fooling, adebayo2022post, bilodeau2024impossibility, wei2024revisiting}. The model must be reliable on the phenomenon, to a degree the relevant scientific community judges adequate; and the explanation must be faithful to the model, to a degree the methodological literature judges adequate. This paper asks a question that neither refinement addresses. \textit{Does a reliable model paired with a faithful explanation of it deliver, by composition alone, a justified claim about the structure of the phenomenon?} We argue that it does not, and that the shortfall is not closed by raising either standard further.

The contribution of the paper is one of clarification. The practice of explaining reliable models invites a move from a faithful account of what the model does to a claim about the structure of the world, and it is this inference we examine. It composes two assessments of different epistemic kinds, a description of what the model computes and an endorsement of the model as a tracker, into a third that is neither, a claim about the structure of the phenomenon. The composition does not go through, and its failure is structural, holding at the limit of perfect reliability and faithfulness as firmly as in any actual case.
 
\section{The chain}
We begin with a formalism familiar to the ML community. Let $f \colon \mathcal{X} \to \mathcal{Y}$ denote the true function representing the natural phenomenon of interest. A predictive model is trained to approximate $f$ by selecting a hypothesis $h^* \in \mathcal{H}$, where $\mathcal{H}$ is a hypothesis space such as the set of functions representable by a deep network of given architecture. We denote $h^*$ a `reliable enough' model\footnote{We adopt the locutions `reliable enough' and `faithful enough' in the spirit of Elgin's `true enough' \citep{elgin2017true}, on which a representation's acceptability is not absolute but depends on whether the larger body of discourse it figures in, such as argument, explanation, or theory.}. For an opaque model $h^*$, a post-hoc explanation method $E$ produces an explanatory representation $g^* = E(h^*) \in \mathcal{P}$, drawn from a space of possible explanations $\mathcal{P}$ such as the SHAP-value space, the saliency-map space, or the counterfactual space. We denote $g^*$ a `faithful enough' explanation in the same contextual sense.
 
The relationship under analysis is the relationship between $f$ and the composition that connects it to $g^*$. A model $h^*$ approximates $f$ through training, and an explanation $g^*$ approximates $h^*$ through some XAI method. We make the composition explicit by writing $f \overset{\phi}{\longleftarrow} h^* \overset{E}{\longleftarrow} g^*$ with arrows running in the direction of approximation: $g^*$ approximates $h^*$, which approximates $f$. We follow Sullivan in calling $\phi$ the \textit{model--world} link \citep{sullivan2024sides}, since it captures whether $h^*$ tracks $f$. For the second link we depart from Sullivan's terminology. Sullivan calls the relation between $h^*$ and $g^*$ a \textit{model--model} link, but the word \textit{model} then covers two different roles. $h^*$ is a model of the phenomenon; $g^*$ is a model of $h^*$, a representation of a representation, one remove further from the world. Naming both relata `model' puts them on a level and so invites the reading that $g^*$, like $h^*$, is a candidate representation of the phenomenon. That reading is precisely what this paper denies. We adopt instead \citeauthor{fleisher2022understanding}'s term and call $E$ the \textit{model-of-model} link, with $g^*$ a model of $h^*$, where the `of' keeps the asymmetry in view.
 
\section{Two verdicts of different epistemic kinds}

The chain $f \leftarrow h^* \leftarrow g^*$ is assembled from two per-link assessments. The argument of this paper is that the two are verdicts of different epistemic kinds, that what the scientist wants is a verdict of a third kind, and that no composition of the first two yields the third. This section sets out the two verdicts and the third thing wanted. $\S4$ shows the composition cannot supply it.

The verdict on $g^*$ as a model of $h^*$ is \textit{descriptive}. A faithful explanation correctly reports which features the model relies on, or more generally what the model computes, in whatever idiom the explanation employs, feature attribution, counterfactual, or concept activation. This is a fact about the computational artefact in $\mathcal{H}$, recovered by the map $E : \mathcal{H} \to \mathcal{P}$. It answers a question about what the model does, and answering it well is what the discipline means by faithfulness. Where faithfulness is checked by comparing recovered features against external annotations, the check confirms the explanation only on the assumption that $h^*$ already tracks $f$, and so tests the descriptive verdict indirectly through the reliability verdict rather than appealing to the world directly.


The verdict on $h^*$ as a tracker of $f$ is partly \textit{justificatory}. When $h^*$ predicts $f$ accurately across the population the surrounding practice has validated against, the endorsement does more than record a regularity in the outputs; it licenses using them as evidence about $f$. A prediction then gives the scientist some reason, evidential and not merely behavioural, to believe the corresponding claim about the phenomenon. This is the standing that reliabilist accounts of model-based belief pick up \citep{duran2023machine}, and that Sullivan's appeal to suitable model-target linking secures \citep{sullivan2022understanding}. Its source is the world, not the model. What licenses belief is the empirical record of the model's tracking against the phenomenon, not any feature of its internal organisation.

What the scientist using XAI on an opaque model wants is a verdict of neither kind. The deliverable is meant to be a justificatory claim about the phenomenon's structure, that some feature, concept, or counterfactual relation recovered from the model corresponds to a structure in the system under study. This is not a description of what the model did, nor an endorsement of the model as a tracker. It is a substantive claim about the world, offered with reasons. By the structure of the phenomenon we mean its causal, mechanistic, or counterfactual organisation, not the merely predictive regularities that reliability and faithfulness may capture. The chain provides a descriptive verdict about the relation between $g^*$ and $h^*$, and a justificatory verdict about $h^*$ as a tracker of $f$. Whether those two can combine into a justificatory verdict about the structure of $f$ is the question of the next section, and what we argue is that they cannot.
 
\section{Why the gap does not close in the limit}

It might be thought that the gap between description-of-model and description-of-world narrows as reliability and faithfulness improve. A perfectly faithful account of a model that perfectly tracks a phenomenon is, on this view, hard to distinguish from an account of the phenomenon itself. The thought is mistaken, and the reason it is mistaken brings out what is structural rather than contingent in the failure.
 
Tracking, however reliable, does not require shared structure. ML models in particular find feature relationships highly divorced from those of their targets, relying on computations or correlations rather than on the causal relationships responsible for the phenomenon \citep{sullivan2024machine}. \citeauthor{sullivan2024machine} takes this divergence to motivate rejecting what she calls the \textit{ML representation hypothesis}: ML models can be epistemically useful without being similar to their targets, provided an interpretative function maps model-facts to claims about the world. The concession is telling, because the epistemic standing that survives the denial of similarity is secured by resources outside the model, such as interpretative functions or embedding in confirmed theory, rather than by reliability of tracking itself. The further fact that would license the inferential move from model to world is not delivered by the verdict of reliability.
 
A reliabilist \citep{duran2023machine} might push back that for certain knowledge claims tracking is sufficient and shared structure is not required. We agree for prediction, where one may be justified in believing the model's output approximately right about $f$ with no further account of why, and disagree only about the claim at issue in scientific XAI. Those claims are about the structure of $f$ in the sense just given, which is to say about mechanism, and mechanism is what shared structure would underwrite. The reliabilist verdict on $\phi$ is right for predictive use; it is the wrong verdict to compose with $E$ when the target is the phenomenon's structure.
 
Faithfulness, however precise, remains a verdict about the model. Improving faithfulness improves the resolution of the description of an element of $\mathcal{H}$ via an element of $\mathcal{P}$ and nothing else. For instance, a circuit-level account of a network is more informative than a feature-attribution account, but it is more informative about the network\footnote{Circuit-level analysis belongs to mechanistic interpretability, which is sometimes set apart from the traditional post-hoc methods rather than counted among them. A circuit-level account, like an attribution, is recovered by a map $E : \mathcal{H} \to \mathcal{P}$; ``faithful description'' here covers any $g^* = E(h^*)$, mechanistic accounts included, and the verdict it yields is descriptive of the model on either reading.}. Whether the operations the network performs correspond to anything in the system the network was trained on is a question that faithful description, on its own, cannot answer. The answer requires premises drawn from \textit{elsewhere}: independent theory about the phenomenon, convergent evidence from systems with known structure, mechanistic hypotheses with prior empirical support, or interventions on the phenomenon itself that confirm or disconfirm the candidate \citep{boge2022two,sullivan2024machine}. These premises do not arrive \textit{with} the chain. They have to be \textit{brought} to it.

Behind these two separate shortfalls lies a single structural reason. Neither verdict ever takes the phenomenon's structure as an input. Reliability is computed from the model's agreement with the phenomenon's outcomes, and faithfulness, since $g^* = E(h^*)$, from the model alone. So what would distinguish a model that has found the phenomenon's structure from one that has merely tracked its outcomes is always a premise brought to the chain from outside, never a tightening of the two verdicts within it. Appendix~\ref{app:mechanism} makes this precise through three cases ordered by how far the model's agreement with the phenomenon reaches.

The claims about the phenomenon that scientific use of the chain aspires to are claims of exactly the sort the chain cannot characterise on its own terms, and improving the verdicts already in hand does not introduce the missing one. Scientific practice itself registers this. \textit{A scientist who has trained a deep network for some scientific prediction task and produced a faithful attribution would not on that basis alone publish a claim that the attributed feature is mechanistically responsible for the phenomenon}. The relevant literature on the feature's plausible role would be consulted, the model's reliance compared against what is known mechanistically in comparable systems, and follow-up experimental or theoretical work undertaken to test the proposed mechanism. What looks like a methodological habit \textit{is} a tacit recognition of the structural deficit. The chain's two verdicts, on their own, do not deliver the verdict the mechanistic claim requires.
 
\section{Strong and weak chains}
The inferential ambitions of scientific XAI vary, and the structural failure has different consequences for different ambitions. Two readings of the chain are worth distinguishing.
 
A \textbf{strong reading} takes the chain to underwrite \textit{how-actually} claims about the phenomenon. The model has learned a function whose internal structure factors through some feature, concept, or counterfactual relation $X$, the explanation faithfully reports that factorisation, and the inference is that $f$ itself factors through $X$. The reading is what makes these claims count as scientific contributions about their targets rather than as descriptions of impressively accurate instruments. The distinction\footnote{\citet{raz2024importance} sharpen the contrast, taking how-actually explanations to identify the real mechanism where how-possibly explanations deliver at most a weaker form of understanding, which they assimilate to objectual understanding.} between \textit{how-possibly} and \textit{how-actually} explanations is owed to Sullivan \citeyearpar{sullivan2022understanding}, who treats how-possibly explanations as specifying possible causes that may fall short of explaining how the phenomenon is in fact caused.
 
A \textbf{weak reading} takes the chain to deliver \textit{how-possibly} claims, candidate hypotheses, or what \citet{baumberger2016understanding} call objectual understanding of the phenomenon. The model offers a possible structure; the explanation describes it; the scientific community is given something worth investigating. Sullivan's positive thesis, that idealised models can deliver how-possibly understanding through suitable linking, defends a version of the weak reading \citep{sullivan2022understanding,sullivan2024machine}. \citet{zednik2021solving} make a similar case for XAI as a tool of scientific exploration rather than of explanation.
 
The structural failure rules out the strong reading as a verdict the chain can supply on the basis of its own two verdicts. Reliability and faithfulness, however ideal, do not license how-actually claims about the phenomenon, because no composition of a justificatory verdict about the model-world link with a descriptive verdict about the model-of-model link supplies the missing premise about shared structure between $h^*$ and $f$. The argument is not that the strong reading is empirically unsupported in present practice, nor that it is in principle unattainable. The argument is that the chain's verdicts are not of the right kinds to compose into the verdict the strong reading requires. Whether external commitments about the phenomenon could supply the missing verdict is a question that lies beyond the scope of the present paper; we set it aside here. 

Yet the weak reading does not rest on the chain's verdicts alone either. A faithful description of a reliable model, considered in isolation from any prior commitments about the phenomenon, is a description of an instrument; nothing in the chain distinguishes the features that bear on $f$'s structure from those that are predictively useful artefacts of the training distribution. To count as a candidate hypothesis about $f$, an item of objectual understanding, or a how-possibly claim worth taking seriously, the description must be situated against prior theory, must converge with evidence from systems whose structure is independently known, or must admit of empirical test against the phenomenon. The difference between the readings is therefore not that one needs external conditions and the other does not. It is that the external conditions \textit{elevate} the chain's output to the status of a candidate.

\section{The philosophical standing of the weak reading}

The grounds on which the weak reading survives have a precise location in the literature on scientific understanding. Three are worth setting out in turn, since each addresses a different feature of what the chain delivers and what it does not.
 
The first ground is the structure of objectual understanding itself. Objectual understanding, as \citet{beisbart2022philosophy} and \citet{raz2024importance} define it following \citet{baumberger2016understanding}, is understanding of a domain of things that typically requires knowledge of the domain and a grasp of connections among items in it; the connections may be merely logical or probabilistic; and there can be a degree of understanding without an actual explanation. A faithful description of a reliable model supplies items in the relevant domain (e.g., the features, concepts, or counterfactuals the model is using) and probabilistic connections among them, namely the model's reliable tracking. That is enough for objectual understanding of $f$ in the weak sense. It is not enough for explanatory understanding of $f$, which on stricter readings of \citet{raz2024importance} requires how-actually rather than how-possibly explanations.
 
The second ground is the literature on idealisation. Sullivan \citeyearpar{sullivan2022understanding} treats DNNs as analogous to highly idealised scientific models and so as deliverers of how-possibly explanations alongside predictive accuracy and heuristic role. The idealisation literature she draws on \citep{elgin2017true,potochnik2019idealization,lawler2022scientific} disagrees on what idealisations are metaphysically but agrees that the epistemic value of an idealisation cannot be cashed out in terms of accuracy or fidelity. The corollary is that falsehood qua falsehood is not what would defeat epistemic value \citep{sullivan2024sides}, which is what allows the weak reading to survive even where strict factivity fails. Idealisations earn their place in science not by representing literally but by affording epistemic access to features of their target that would otherwise be difficult or impossible to discern.
 
The third ground concerns the kinds of understanding that the literature distinguishes when assessing DNNs. \citet{paez2020pragmatic} treats understanding-why as a localised variety of objectual understanding in machine learning, and accepts that factivity for understanding-why is doubtful once idealisations enter. The implication for the chain is that the strong reading, which would aspire to factive understanding-why of $f$, is under independent pressure from the philosophical literature even before our structural argument is brought to bear. Functional understanding, in the Lombrozo--Wilkenfeld (\citeyear{grimm2019new}) sense, tells us what something is supposed to do without telling us how it does it, and is the kind of understanding that purely model-neutral interpretability methods are best suited to deliver about the model. Functional understanding is not what scientific XAI aspires to about $f$; if it were, the chain's failure to deliver more would be no failure at all. Objectual understanding is the kind that the weak reading defensibly delivers about $f$, under the conditions just noted. Explanatory understanding of $f$ is the kind the strong reading would deliver and the kind the chain cannot. The factivity the chain can supply is therefore exactly the factivity the weak reading requires, and no more.
 
 
\section{Remarks}

The chain meets some of scientific XAI's aims and not others. Reliability tells the scientist that the model predicts the phenomenon's outcomes; faithfulness tells the scientist what the model relies on in doing so; neither tells the scientist whether the model arrives at those outcomes as the phenomenon does. The chain can therefore support candidate hypotheses under external corroboration, but it cannot, on its own, license claims about how the phenomenon is in fact structured. What would license such claims, including what good corroboration consists in, is a question this paper leaves open, and the right one to ask next.

\bibliography{example_paper}

@article{duran2023machine,
  title={Machine learning, justification, and computational reliabilism},
  author={Dur{\'a}n, Juan Manuel},
  year={2023},
  publisher={Unpublished manuscript}
}

@article{hooker2019benchmark,
  title={A benchmark for interpretability methods in deep neural networks},
  author={Hooker, Sara and Erhan, Dumitru and Kindermans, Pieter-Jan and Kim, Been},
  journal={Advances in neural information processing systems},
  volume={32},
  year={2019}
}

@inproceedings{adebayo2022post,
  title={Post hoc explanations may be ineffective for detecting unknown spurious correlation},
  author={Adebayo, Julius and Muelly, Michael and Abelson, Harold and Kim, Been},
  booktitle={International conference on learning representations},
  year={2022}
}

@article{bilodeau2024impossibility,
  title={Impossibility theorems for feature attribution},
  author={Bilodeau, Blair and Jaques, Natasha and Koh, Pang Wei and Kim, Been},
  journal={Proceedings of the National Academy of Sciences},
  volume={121},
  number={2},
  pages={e2304406120},
  year={2024},
  publisher={National Acad Sciences}
}

@article{zednik2021solving,
  title={Solving the black box problem: A normative framework for explainable artificial intelligence},
  author={Zednik, Carlos},
  journal={Philosophy \& technology},
  volume={34},
  number={2},
  pages={265--288},
  year={2021},
  publisher={Springer}
}

@article{sullivan2022understanding,
  title={Understanding from machine learning models},
  author={Sullivan, Emily},
  journal={The British Journal for the Philosophy of Science},
  year={2022},
  publisher={The University of Chicago Press}
}

@article{beisbart2022philosophy,
  title={Philosophy of science at sea: Clarifying the interpretability of machine learning},
  author={Beisbart, Claus and R{\"a}z, Tim},
  journal={Philosophy Compass},
  volume={17},
  number={6},
  pages={e12830},
  year={2022},
  publisher={Wiley Online Library}
}

@article{laugel2019issues,
  title={Issues with post-hoc counterfactual explanations: a discussion},
  author={Laugel, Thibault and Lesot, Marie-Jeanne and Marsala, Christophe and Detyniecki, Marcin},
  journal={arXiv preprint arXiv:1906.04774},
  year={2019}
}

@article{wei2024revisiting,
  title={Revisiting the robustness of post-hoc interpretability methods},
  author={Wei, Jiawen and Turb{\'e}, Hugues and Mengaldo, Gianmarco},
  journal={arXiv preprint arXiv:2407.19683},
  year={2024}
}

@article{laugel2019dangers,
  title={The dangers of post-hoc interpretability: Unjustified counterfactual explanations},
  author={Laugel, Thibault and Lesot, Marie-Jeanne and Marsala, Christophe and Renard, Xavier and Detyniecki, Marcin},
  journal={arXiv preprint arXiv:1907.09294},
  year={2019}
}

@inproceedings{slack2020fooling,
  title={Fooling lime and shap: Adversarial attacks on post hoc explanation methods},
  author={Slack, Dylan and Hilgard, Sophie and Jia, Emily and Singh, Sameer and Lakkaraju, Himabindu},
  booktitle={Proceedings of the AAAI/ACM Conference on AI, Ethics, and Society},
  pages={180--186},
  year={2020}
}

@article{sullivan2024machine,
title={Do machine learning models represent their targets?},
author={Sullivan, Emily},
journal={Philosophy of Science},
volume={91},
number={5},
pages={1445--1455},
year={2024},
publisher={Cambridge University Press}
}

@inproceedings{sullivan2024sides,
title={SIDEs: Separating Idealization from Deceptive'Explanations' in xAI},
author={Sullivan, Emily},
booktitle={Proceedings of the 2024 ACM Conference on Fairness, Accountability, and Transparency},
pages={1714--1724},
year={2024}
}

@article{fleisher2022understanding,
title={Understanding, idealization, and explainable AI},
author={Fleisher, Will},
journal={Episteme},
volume={19},
number={4},
pages={534--560},
year={2022},
publisher={Cambridge University Press}
}

@article{paez2020pragmatic,
title={The pragmatic turn in explainable artificial intelligence (XAI)},
author={P{\'a}ez, Andr{\'e}s},
journal={arXiv preprint arXiv:2002.09595},
year={2020}
}

@article{boge2022two,
title={Two dimensions of opacity and the deep learning predicament},
author={Boge, Florian J},
journal={Minds and Machines},
volume={32},
number={1},
pages={43--75},
year={2022},
publisher={Springer}
}

@article{raz2024importance,
title={The importance of understanding deep learning},
author={R{\"a}z, Tim and Beisbart, Claus},
journal={Erkenntnis},
volume={89},
number={5},
pages={1823--1840},
year={2024},
publisher={Springer}
}

@book{elgin2017true,
title={True enough},
author={Elgin, Catherine Z},
year={2017},
publisher={MIT press}
}

@book{potochnik2019idealization,
title={Idealization and the Aims of Science},
author={Potochnik, Angela},
year={2019},
publisher={University of Chicago Press}
}

@book{lawler2022scientific,
title={Scientific Understanding and Representation: modeling in the physical sciences},
author={Lawler, Insa and Khalifa, Kareem and Shech, Elay},
year={2022},
publisher={Taylor \& Francis}
}

@article{benslimane2024shap,
title={A SHAP-based controversy analysis through communities on Twitter},
author={Benslimane, Samy and Papastergiou, Thomas and Az{\'e}, J{\'e}r{\^o}me and Bringay, Sandra and Servajean, Maximilien and Mollevi, Caroline},
journal={World Wide Web},
volume={27},
number={5},
pages={65},
year={2024},
publisher={Springer}
}

@article{auletta2023predicting,
title={Predicting and understanding human action decisions during skillful joint-action using supervised machine learning and explainable-AI},
author={Auletta, Fabrizia and Kallen, Rachel W and Di Bernardo, Mario and Richardson, Michael J},
journal={Scientific Reports},
volume={13},
number={1},
pages={4992},
year={2023},
publisher={Nature Publishing Group UK London}
}

@article{kumar2024shared,
title={Shared functional specialization in transformer-based language models and the human brain},
author={Kumar, Sreejan and Sumers, Theodore R and Yamakoshi, Takateru and Goldstein, Ariel and Hasson, Uri and Norman, Kenneth A and Griffiths, Thomas L and Hawkins, Robert D and Nastase, Samuel A},
journal={Nature communications},
volume={15},
number={1},
pages={5523},
year={2024},
publisher={Nature Publishing Group UK London}
}

@article{wang2024rethink,
title={Rethink data-driven human behavior prediction: A Psychology-powered Explainable Neural Network},
author={Wang, Jiyao and Huang, Chunxi and Xie, Weiyin and He, Dengbo and Tu, Ran},
journal={Computers in Human Behavior},
volume={156},
pages={108245},
year={2024},
publisher={Elsevier}
}

@article{ji2026multi,
title={Multi-Scale Explainable AI for RMB Exchange Rate Drivers},
author={Ji, Jie and Wang, Shouyang and Wei, Yunjie},
journal={Forecasting},
volume={8},
number={1},
pages={7},
year={2026},
publisher={MDPI}
}

@article{chung2023inside,
title={Inside the black box: Neural network-based real-time prediction of US recessions},
author={Chung, Seulki},
journal={arXiv preprint arXiv:2310.17571},
year={2023}
}

@article{yuan2024deep,
title={Deep learning interpretability for rough volatility},
author={Yuan, Bo and Brigo, Damiano and Jacquier, Antoine and Pede, Nicola},
journal={arXiv preprint arXiv:2411.19317},
year={2024}
}

@article{goswami2026significance,
title={Significance of predictors: revisiting stock return predictions using explainable AI},
author={Goswami, Bhaskar and Uddin, Ajim},
journal={Annals of Operations Research},
volume={357},
number={1},
pages={223--257},
year={2026},
publisher={Springer}
}

@article{eilts2026explainable,
title={Explainable AI Insights Into EEG Classification and Its Alignment to Neural Correlates},
author={Eilts, Hendrik and Ivucic, Gabriel and Koenen, Niklas and Wright, Marvin N and Schultz, Tanja and Putze, Felix},
journal={Human Brain Mapping},
volume={47},
number={6},
pages={e70528},
year={2026},
publisher={Wiley Online Library}
}

@article{goktepe2025does,
title={What does my network learn? Assessing interpretability of deep learning for EEG},
author={G{\"o}ktepe-Kavis, Pinar and Aellen, Florence M and Alnes, Sigurd L and Tzovara, Athina},
journal={Imaging Neuroscience},
year={2025}
}

@article{ryali2024deep,
title={Deep learning models reveal replicable, generalizable, and behaviorally relevant sex differences in human functional brain organization},
author={Ryali, Srikanth and Zhang, Yuan and de Los Angeles, Carlo and Supekar, Kaustubh and Menon, Vinod},
journal={Proceedings of the National Academy of Sciences},
volume={121},
number={9},
pages={e2310012121},
year={2024},
publisher={National Academy of Sciences}
}

@article{hofmann2025utility,
title={The utility of explainable AI for MRI analysis: Relating model predictions to neuroimaging features of the aging brain},
author={Hofmann, Simon M and Goltermann, Ole and Scherf, Nico and M{\"u}ller, Klaus-Robert and L{\"o}ffler, Markus and Villringer, Arno and Gaebler, Michael and Witte, A Veronica and Beyer, Frauke},
journal={Imaging Neuroscience},
volume={3},
pages={imag\_a\_00497},
year={2025},
publisher={MIT Press 255 Main Street, 9th Floor, Cambridge, Massachusetts 02142, USA~…}
}

@article{vinokic2025effectiveness,
title={Effectiveness of three machine learning models for prediction of daily streamflow and uncertainty assessment},
author={Vinoki{\'c}, Luka and Dotli{\'c}, Milan and Prodanovi{\'c}, Veljko and Kolakovi{\'c}, Slobodan and Simonovic, Slobodan P and Stojkovi{\'c}, Milan},
journal={Water Research X},
volume={27},
pages={100297},
year={2025},
publisher={Elsevier}
}

@article{vent2026physics,
title={The physics behind ML-based quark-gluon taggers},
author={Vent, Sophia and Winterhalder, Ramon and Plehn, Tilman},
journal={SciPost Physics},
volume={20},
number={3},
pages={084},
year={2026}
}

@article{bonalumi2024explainable,
title={eXplainable artificial intelligence applied to algorithms for disruption prediction in tokamak devices},
author={Bonalumi, Luca and Aymerich, Enrico and Alessi, Edoardo and Cannas, Barbara and Fanni, Alessandra and Lazzaro, Enzo and Nowak, Silvana and Pisano, Fabio and Sias, Giuliana and Sozzi, Carlo},
journal={Frontiers in Physics},
volume={12},
pages={1359656},
year={2024},
publisher={Frontiers Media SA}
}

@article{pampari2025chrombpnet,
title={ChromBPNet: bias factorized, base-resolution deep learning models of chromatin accessibility reveal cis-regulatory sequence syntax, transcription factor footprints and regulatory variants},
author={Pampari, Anusri and Shcherbina, Anna and Kvon, Evgeny Z and Kosicki, Michael and Nair, Surag and Kundu, Soumya and Kathiria, Arwa S and Risca, Viviana I and Kuningas, Kristiina and Alasoo, Kaur and others},
journal={BioRxiv},
pages={2024--12},
year={2025}
}

@article{mao2024phenotype,
title={Phenotype prediction from single-cell RNA-seq data using attention-based neural networks},
author={Mao, Yuzhen and Lin, Yen-Yi and Wong, Nelson KY and Volik, Stanislav and Sar, Funda and Collins, Colin and Ester, Martin},
journal={Bioinformatics},
volume={40},
number={2},
pages={btae067},
year={2024},
publisher={Oxford University Press}
}

@article{sun2025esm2_amp,
title={ESM2\_AMP: an interpretable framework for protein--protein interactions prediction and biological mechanism discovery},
author={Sun, Yawen and Wang, Rui and Luo, Zeyu and Tan, Lejia and Liu, Junhao and Li, Ruimeng and Wei, Dongqing and Zhang, Yu-Juan},
journal={Briefings in Bioinformatics},
volume={26},
number={4},
pages={bbaf434},
year={2025},
publisher={Oxford University Press}
}

@article{klein2026explainable,
title={Explainable AI-based analysis of human pancreas sections identifies traits of type 2 diabetes},
author={Klein, Lukas and Ziegler, Sebastian and Gerst, Felicia and Morgenroth, Yanni and Gotkowski, Karol and Sch{\"o}niger, Eyke and Heni, Martin and Kipke, Nicole and Friedland, Daniela and Seiler, Annika and others},
journal={Nature Communications},
year={2026},
publisher={Nature Publishing Group UK London}
}

@article{atad2022chexplaining,
title={Chexplaining in style: Counterfactual explanations for chest x-rays using stylegan},
author={Atad, Matan and Dmytrenko, Vitalii and Li, Yitong and Zhang, Xinyue and Keicher, Matthias and Kirschke, Jan and Wiestler, Bene and Khakzar, Ashkan and Navab, Nassir},
journal={arXiv preprint arXiv:2207.07553},
year={2022}
}

@article{rahnfeld2025comparative,
title={A comparative study of explainability methods for whole slide classification of lymph node metastases using vision transformers},
author={Rahnfeld, Jens and Naouar, Mehdi and Kalweit, Gabriel and Boedecker, Joschka and Dubruc, Estelle and Kalweit, Maria},
journal={PLOS digital health},
volume={4},
number={4},
pages={e0000792},
year={2025},
publisher={Public Library of Science San Francisco, CA USA}
}

@article{stomberg2023recognizing,
title={Recognizing protected and anthropogenic patterns in landscapes using interpretable machine learning and satellite imagery},
author={Stomberg, Timo T and Leonhardt, Johannes and Weber, Immanuel and Roscher, Ribana},
journal={Frontiers in Artificial Intelligence},
volume={6},
pages={1278118},
year={2023},
publisher={Frontiers Media SA}
}

@article{graciosa2025uncovering,
title={Uncovering deformation prior to analogue megathrust earthquakes with Explainable Artificial Intelligence},
author={Graciosa, Juan Carlos and Corbi, Fabio and Capitanio, Fabio A},
journal={Geophysical Research Letters},
volume={52},
number={12},
pages={e2024GL114428},
year={2025},
publisher={Wiley Online Library}
}

@article{baumberger2016understanding,
  title={What is understanding? An overview of recent debates in epistemology and philosophy of science},
  author={Baumberger, Christoph and Beisbart, Claus and Brun, Georg},
  journal={Explaining understanding},
  pages={1--34},
  year={2016},
  publisher={Routledge}
}

@article{grimm2019new,
  title={New Perspectives from Philosophy, Psychology, and Theology},
  author={GRIMM, STEPHEN R},
  year={2019}
}

@article{degrave2021ai,
  title={AI for radiographic COVID-19 detection selects shortcuts over signal},
  author={DeGrave, Alex J and Janizek, Joseph D and Lee, Su-In},
  journal={Nature Machine Intelligence},
  volume={3},
  number={7},
  pages={610--619},
  year={2021},
  publisher={Nature Publishing Group UK London}
}
\bibliographystyle{icml2026}

\newpage
\appendix
\onecolumn

\section{Why the gap does not close, in three cases}
\label{app:mechanism}

Section~4 claims that neither of the chain's verdicts registers the phenomenon's structure, and that the resulting gap survives however far reliability and faithfulness are pushed. This appendix establishes the claim by working through the three cases, ordered by how far the model's agreement with the phenomenon extends, and then develops the single route by which all three mislead the scientist. The first is a confound that fails outside the training distribution. The second is a confound that holds across every population the practice can test on. The third is a model that agrees with the phenomenon everywhere. 

We begin with the case easiest to catch. \citet{degrave2021ai} trained deep networks to detect a viral respiratory infection from chest radiographs, and the networks predicted well on held-out data while saliency methods faithfully reported which features drove them. Some of the regions the maps flagged were not pathology but laterality markers, the image edges, and the cardiac silhouette. That these regions actually drove the predictions was not assumed but tested by intervention, since swapping a laterality marker between a positive and a negative image moved the output far more than swapping a random patch of the same size. These features predicted the diagnosis only because the positive and negative radiographs had been gathered from different sources, so that source differences coincided with label differences. A coincidence of that kind is specific to the training sources, so the reliance ought to break wherever the images come from elsewhere, and it did. Tested on new hospitals, where the coincidence no longer held, the networks lost about half their performance. Throughout, the explanation stayed faithful, reporting the reliance on laterality markers in the same idiom it used for lung opacity, and nothing in that idiom marked one feature as spurious and the other as genuine. So what separated them was knowledge of where the images came from and of what radiologists examine, which is knowledge about the phenomenon and not about the model.

The second case is one that no amount of validation can catch. Suppose the spurious feature is correlated with the target not only in the training data but across every population the practice is able to test. A potential example is patient positioning, the way a body is arranged for the image. Only the part of positioning that does not follow from the radiographic projection counts as the confound, and that part keeps a consistent relationship with the label across both of the study's datasets. External validation removes only features that vary across populations, and this one does not vary. So a model can read positioning in place of pathology, predict the diagnosis everywhere the practice looks, pass external validation at every turn, and still rely on a feature that is not pathology \citep{degrave2021ai}.

To see why testing cannot expose such a feature, recall what a validation check is. It runs the model on some data and records how closely the predictions match the outcomes, and the result it returns summarises those predictions and contains nothing else. Now place two models side by side, one reading the genuine feature and one reading the proxy. On every population available for testing both features agree with the label, and if two features track the label identically then models built on them produce identical predictions, so across all of that data the two models predict alike. Identical predictions produce identical validation records, and since the result just described contains nothing but those records, no check built on them can tell the two models apart.

The natural response is to test on more populations, and it repays seeing exactly why this does not help. Adding a population yields one of two outcomes. In the first, the two features disagree there, so the proxy fails, performance drops, and the problem becomes visible, but that is the first case over again and not the one now under consideration. In the second, the features still agree, so the two models once more predict alike and the record is no more telling than before. By the definition of the second case the proxy holds across everything the practice can reach, so only the second outcome is available, and adding data therefore removes whichever proxies happen to break and leaves the rest exactly where they stood.

What the extra data could never contain, a single patient would settle. Take a patient who is genuinely sick but arranged on the table as a healthy patient would be. He is the one case on which reading pathology and reading positioning predict different diagnoses, since the disease points one way and the posture the other, so his diagnosis would reveal at a stroke which of the two the model had been relying on. The difficulty is that the second case is defined by sickness and healthy-style positioning never occurring together in any patient one can simply collect, so the deciding patient never arises of his own accord however many are gathered, which is exactly why more data was powerless. He has instead to be produced on purpose, and there are two ways to do it. One alters the image directly, introducing the pathology while holding the positioning fixed. The other seeks out a population where the correlation is known on independent grounds to break (e.g., a ward where every patient, sick or healthy, is imaged under a single fixed positioning protocol, so that positioning no longer follows the diagnosis). Both require a prior commitment about the phenomenon, either which manipulation changes the pathology or where the correlation gives way, and the chain supplies no such commitment.

The third case removes the confound entirely and shows that even then the gap remains. Suppose the model agrees with the phenomenon on every input, not merely on the populations available for testing, and leans on no spurious feature. This is the perfect tracker that the objection at the head of Section~4 had in mind. The claim is that the gap survives because agreement in behaviour fixes nothing about agreement in structure, as a pair of models shows. The first computes the phenomenon's input-output map through internal structure matching the phenomenon's own; the second computes the very same map through internal structure of its own. Both agree with the phenomenon everywhere, so both are perfect trackers, yet their internal structures differ, and this is no contrived pairing, since one pattern of behaviour can be realised by many mechanisms and tracking a target through structure unlike its own is the ordinary case in ML \citep{sullivan2024machine}. Now ask what reliability and faithfulness report of the two models. Reliability asks only whether the model's outputs match the phenomenon's outputs, and that match is perfect for each model, so reliability comes out the same for both. Faithfulness asks only whether $g^*$ correctly describes $h^*$, and because $g^* = E(h^*)$ it is computed from the model itself, drawing at most on the same record of inputs and outputs and never on how the phenomenon is internally built, so faithfulness too comes out the same for both. Neither measure takes the phenomenon's structure as an input, so neither can register the one respect in which the models differ, and both come out identical for the structure-matching model and the structure-diverging one.

It might seem that the structure-matching model rescues the chain. Because its internals mirror the phenomenon, a faithful description of it reports the phenomenon's real structure, and a scientist who believes that report believes something true. Grant that the report is true; the trouble is not its truth but what entitles her to believe it. All she has to go on is that her model tracks the phenomenon perfectly and that its explanation faithfully reports how the model is organised, and those same two things would hold of her in either of two situations. In the first, her model is the structure-matching one, its explanation reports the phenomenon's real structure, and her belief is true. In the second, her model is the structure-diverging one, which tracks the phenomenon just as perfectly through a structure of its own, its explanation faithfully reports that other structure, and a scientist trusting it in the same way believes a falsehood. Reliability and faithfulness are good enough in both situations, so they are satisfied whether the reported structure is the phenomenon's or not, and what separates the two situations, namely whether the model's structure is the phenomenon's, is exactly what the two measures never check. Credentials equally satisfied when the belief is false cannot be what justifies it when true. So in the first situation the belief is true, but its truth is not owed to reliability and faithfulness, since a scientist equally backed by both believes a falsehood in the second. The belief is true because she was handed the structure-matching model rather than the other, which is luck rather than anything her credentials revealed, and a belief true only by such luck is not knowledge. The point is at bottom one about composition. Both measures are fully met by the structure-diverging model, whose structure is wrong, so no rule that combines them, however weighted or thresholded, can single out the model whose structure is right. The truth of a structural description is thus a feature the description may have but that reliability and faithfulness cannot establish, since both hold equally when it is false, and perfect reliability and perfect faithfulness together leave the structural question exactly where it stood.

One reply deserves direct treatment, since it appears to escape the argument. The three cases turn on reliability and faithfulness never taking the phenomenon's structure as an input, so suppose we feed that structure in by another route, constraining the hypothesis space with causal assumptions so that only models built to factor through the phenomenon's real organisation are allowed. A faithful explanation of such a model would then seem to describe the phenomenon's structure after all, since the model was built to share it. Grant that it does, that a constrained model can yield a world-directed explanation; the question is what produced that result. The causal constraint is itself a claim about the phenomenon, asserted when the architecture is chosen, so it is the same external premise the three cases found the chain unable to supply, only introduced earlier, at the choice of model rather than after training. Where a constrained chain seems to report the phenomenon's structure, that structure was placed into the hypothesis space by assumption, not drawn out of the model by reliability and faithfulness. Structure can enter through the data (e.g., curating the training set so that a known confound such as image source no longer tracks the diagnosis), through the architecture (e.g., building a known constraint on the form of the solution into the model so that it can only represent mechanisms of the intended kind), or through the interpretation laid over the output (e.g., reading a recovered feature as mechanistic only where independent theory already says it could be), but never through reliability and faithfulness themselves.

Consider next what the constraint actually pins down. It fixes the model at one grain, the grain of which parts the computation has and how they feed into one another, and at that grain the model is forced to match the phenomenon, so a faithful description recovers a structure that is genuinely the phenomenon's. But the constraint says nothing about the grain beneath, namely how the model carries out each of those steps, and there the model is left free to use whatever internal machinery fits the data. The phenomenon carries out the same steps by some detailed means of its own, and the model's means need not resemble them at all. So beneath the grain the constraint reaches, the model and the phenomenon can be built differently while still behaving identically, which is precisely the third case again, now reappearing one grain down inside a model that was supposed to have been constrained. The only way to shut this down is to write the constraint all the way to the bottom, fixing not just the parts and their connections but every detail of how each step is carried out. Yet a mechanism specified to the bottom by hand has been put there by the modeller and not learned from the model, so it has been brought from outside and discovered nowhere, just as before. The constraint thus does two different things at two grains. At the grain it reaches it defeats the confounds, since a model forced through the right parts cannot substitute positioning for pathology. Beneath that grain it does nothing, and the difference of realisation from the third case survives untouched.

The three cases share a single route to misleading the scientist. In each, the explanation does its job and reports accurately what the model relies on. What it cannot do is mark whether the feature it reports belongs to the phenomenon, and it cannot, because that is a question about the phenomenon and the explanation sees only the model. A scientist who reads the explanation as a window onto the phenomenon takes the reported feature to be a feature of $f$, and so consumes a description of the model as though it were a justification about the world, which is the composition Section~3 finds ill-typed. 

\end{document}